\documentclass[10pt,twocolumn,letterpaper]{article}

\usepackage{iccv}
\usepackage{times}
\usepackage{epsfig}
\usepackage{graphicx}

\usepackage{amsmath,amssymb,amsfonts}
\usepackage{algorithmic}
\usepackage[algo2e]{algorithm2e} 

\usepackage{textcomp}

\usepackage[inline]{enumitem}
\usepackage{xcolor}
\usepackage{caption}

\usepackage{algorithm}  
\usepackage{subfigure}
\usepackage[T1]{fontenc}
\usepackage{aecompl}

\usepackage{diagbox,slashbox}
\usepackage{verbatim}
\usepackage{multirow} 
\usepackage{multicol} 
\usepackage{tabularx}
\usepackage{hhline}

\usepackage[pagebackref=true,breaklinks=true,letterpaper=true,colorlinks,bookmarks=false]{hyperref}

 \iccvfinalcopy 

\def\iccvPaperID{11217} 

\ificcvfinal\fi

\begin{document}

\title{Semi-supervised Semantics-guided Adversarial
Training for \\ Robust Trajectory Prediction}

\author{Ruochen Jiao\\
Northwestern University\\
{\tt\small ruochen.jiao@u.northwestern.edu}
\and
Xiangguo Liu\\
Northwestern University\\
{\tt\small xg.liu@u.northwestern.edu}
\and
Takami Sato\\
 University of California, Irvine\\
{\tt\small takamis@uci.edu}
\and
Qi Alfred Chen\\
 University of California, Irvine\\
{\tt\small alfchen@uci.edu}
\and
Qi Zhu\\
Northwestern University\\
{\tt\small qzhu@northwestern.edu}
}

\maketitle
\ificcvfinal\thispagestyle{empty}\fi

\begin{abstract}
Predicting the trajectories of surrounding objects is a critical task for self-driving vehicles and many other autonomous systems. 
Recent works demonstrate that adversarial attacks on trajectory prediction, where small crafted perturbations are introduced to history trajectories, may significantly mislead the prediction of future trajectories and induce unsafe planning. However, few works have addressed enhancing the robustness of this important safety-critical task. In this paper, we present a \textbf{novel adversarial training method for trajectory prediction}. 
Compared with typical adversarial training on image tasks, our work is challenged by more random input with rich context and a lack of class labels. To address these challenges, we propose a method based on a \textbf{semi-supervised} adversarial autoencoder, which models \textbf{disentangled semantic features} with domain knowledge and provides additional latent labels for the adversarial training. 
 Extensive experiments with different types of attacks demonstrate that our \textbf{S}emi-supervised \textbf{S}emantics-guided \textbf{A}dversarial \textbf{T}raining  (\emph{\textbf{SSAT}}) method can effectively mitigate the impact of adversarial attacks by up to $73\%$ and outperform other popular defense methods. In addition, experiments show that our method can significantly improve the system’s robust generalization to unseen patterns of attacks. We believe that such semantics-guided architecture and advancement on robust generalization is an important step for developing robust prediction models and enabling safe decision making.
\end{abstract}

\section{Introduction}
The adversarial robustness of deep neural networks (DNNs) has drawn significant attention in recent years. State-of-the-art classifiers can be misled to make erroneous predictions by slight but carefully-optimized perturbations~\cite{szegedy2013intriguing}.
Besides classification, adversarial attacks are also targeted at various other tasks such as object detection, image generation and natural language processing. To defend against adversarial examples, adversarial training is commonly used to enhance the intrinsic robustness of models and has been shown to be very effective among various defense strategies~\cite{madry2017towards,maini2020adversarial,schott2018towards}. On the other hand, the works in~\cite{rice2020overfitting,wu2020adversarial,song2019robust} show that adversarial training may suffer poor robust generalization on unseen attacks and~\cite{schmidt2018adversarially,madry2017towards,su2018robustness} demonstrate a trade-off between robustness and accuracy. In this paper, we address \textbf{trajectory prediction}, a safety-critical task that is common in self-driving vehicles and other autonomous systems, and propose a novel architecture to enhance its \emph{adversarial robustness and robust generalization} by introducing semantic features and a semi-supervised adversarial autoencoder (AAE)~\cite{makhzani2015adversarial} into the adversarial training process. 

\begin{figure}[ht]
    \centering
    \includegraphics[width=\columnwidth]{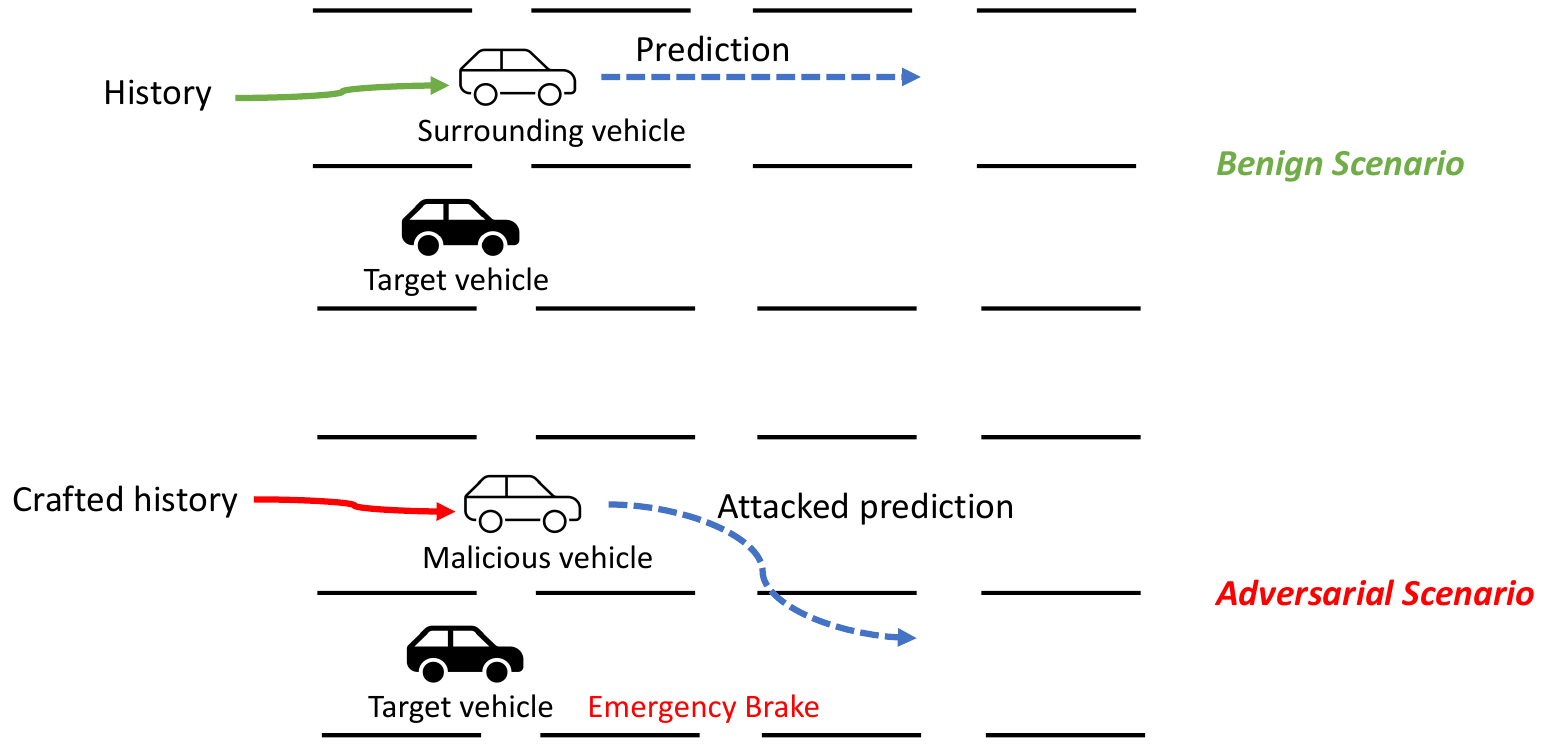}
    \caption{An illustrative example of the threat model: In the upper benign scenario, the target vehicle predicts the future trajectory of a surrounding vehicle based on its history trajectory. In the lower adversarial scenario, the surrounding vehicle is malicious and drives along a crafted (history) trajectory, which may mislead the target vehicle to have a wrong (attacked) prediction of the surrounding vehicle's future trajectory. In this particular case, the target vehicle wrongly predicts that the malicious vehicle will cut into its lane, and thus takes an unsafe emergency bake. }
    \label{fig:intro}
\end{figure}

In particular, we focus on the trajectory prediction module in autonomous driving, where large DNNs have enabled breakthroughs in recent years. More specifically, an autonomous driving system typically consists of several modules such as perception, localization, prediction, planning (route, behavior, trajectory, and motion planning), and control. The perception module is to detect agents and obstacles based on the sensing inputs (e.g., images and 3D point clouds), and the effectiveness and robustness of the perception module are well-studied in the research community. The prediction module is to predict \emph{future trajectories} of surrounding vehicles based on the observed \emph{history trajectories} of those vehicles from the perception module and the map context. The trajectory prediction plays a crucial role for understanding the environment and making safety-critical decisions in the following planning module. Recent works~\cite{liu2021multimodal,liang2020learning,gao2020vectornet,yuan2021agentformer,jiao2022tae} have applied various deep learning techniques such as graph neural networks and transformers to this task and achieved impressive performance in terms of reducing average errors. 
However, few works studied the robustness of vehicle trajectory prediction under adversarial attacks, which is in fact of vital importance because 1) autonomous driving is a safety-critical task by nature, 2) recent work~\cite{zhang2022adversarial} demonstrates that the prediction module is vulnerable to adversarial attacks if surrounding vehicles drive along a seemingly natural but crafted trajectory, and 3) current prediction models are often overfitted on limited patterns in the datasets but suffer the long-tailed distribution of driving scenarios and behaviors. The threat model of the adversarial trajectory is illustrated in an example in Fig.~\ref{fig:intro}, where a crafted input trajectory can intentionally mislead the target vehicle to the wrong prediction and induce dangerous planning decisions.

To defend against adversarial attacks on vehicle trajectory prediction, there are important challenges that are different from the cases of images and audios. First, vehicle trajectory prediction is a time-series regression problem with rich contexts while most existing adversarial attacks and corresponding defense methods are targeted at classification tasks. The attack patterns are more random and there are no well-defined class labels, which means that the robust model is difficult to train and generalize. Second, vehicle trajectory can convey semantic and behavior information. For instance, people can infer the behavior of a vehicle such as moving forward or changing lanes from its trajectory. Therefore, as shown in our experiments later, some popular defense methods such as TRADES~\cite{zhang2019theoretically} and data augmentation methods \cite{gowal2021improving,rebuffi2021fixing,sehwag2021robust}  either are inapplicable or have poor performance in trajectory prediction task.  
In this work, we first propose an adversarial training pipeline for the trajectory prediction task, and then further exploit the semantic features to design a semi-supervised AAE architecture that can be added after the feature extractor, to improve the adversarial robustness and its generalization. The methodology of the proposed architecture such as enhancing disentanglement and defining semantic labels may be further applied on adversarial training for general regression and generation problems.
Our main contributions are summarized as follows:
\begin{itemize}
    \item We propose a novel adversarial training method against adversarial attacks on trajectory predictions.
    \item We develop a semi-supervised architecture with domain knowledge and semantic features to enhance the adversarial robustness and its generalization among different types of attack patterns.
    \item Extensive experiments demonstrate that our proposed method (SSAT) effectively improves the adversarial robustness and outperforms popular defense baselines.
    \item We further explore to balance the robustness-accuracy trade-off in this task by leveraging the MixUp technique~\cite{zhang2017mixup}, and we also test data augmentation methods for trajectory prediction robustness.
\end{itemize}

\section{Preliminaries}

\label{related-work}

%

\subsection{Adversarial Attacks on Trajectory Prediction}
\label{sec:pre-att}
Recent research~\cite{zhang2022adversarial} demonstrates that trajectory prediction in autonomous driving can be fooled by the adversarial behavior of a surrounding vehicle, 
where the adversarial behavior is optimized with Projected Gradient Decent (PGD)~\cite{madry2017towards}. 
The threat model is as in Fig.~\ref{fig:intro}. In this work, we consider such attacks and assume the worst setting, i.e., the attacker has full knowledge of the target system and tries to maximize the attack impact.

\begin{figure*}[htbp]
\centering
\label{fig:attack-sample}
\subfigure[]{
\begin{minipage}[t]{0.22\linewidth}
\centering
\includegraphics[width=1.55in]{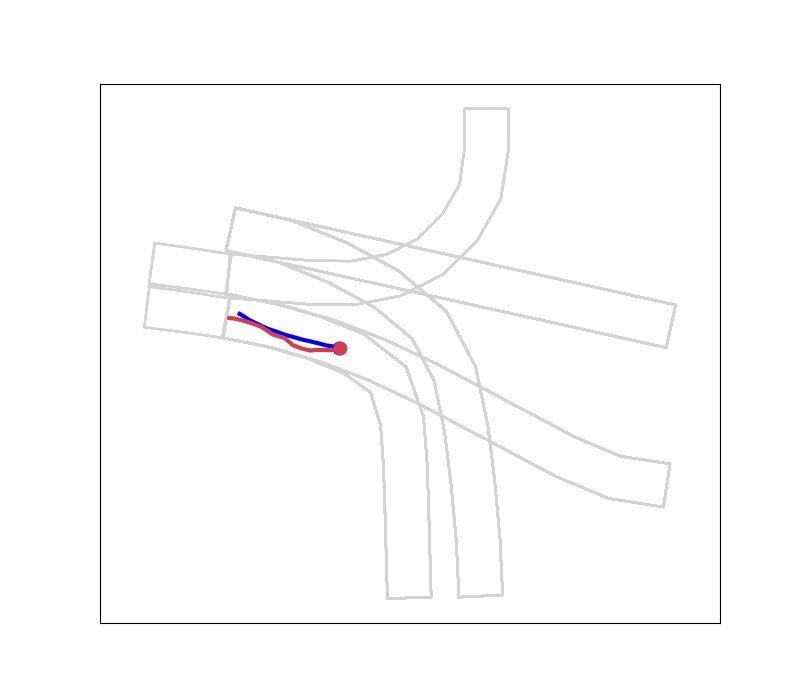}
\\
\includegraphics[width=1.55in]{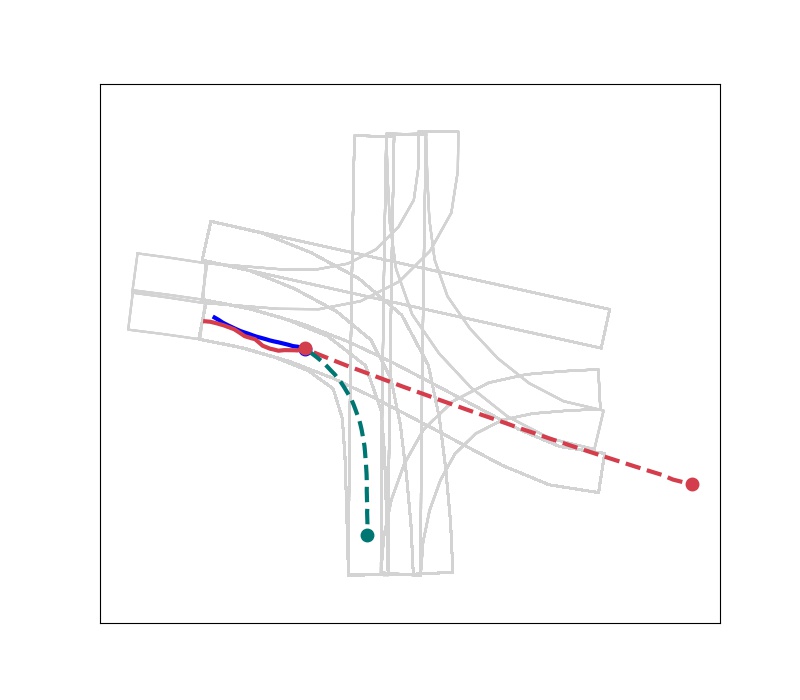}
\end{minipage}%
}%
\subfigure[]{
\begin{minipage}[t]{0.22\linewidth}
\centering
\includegraphics[width=1.55in]{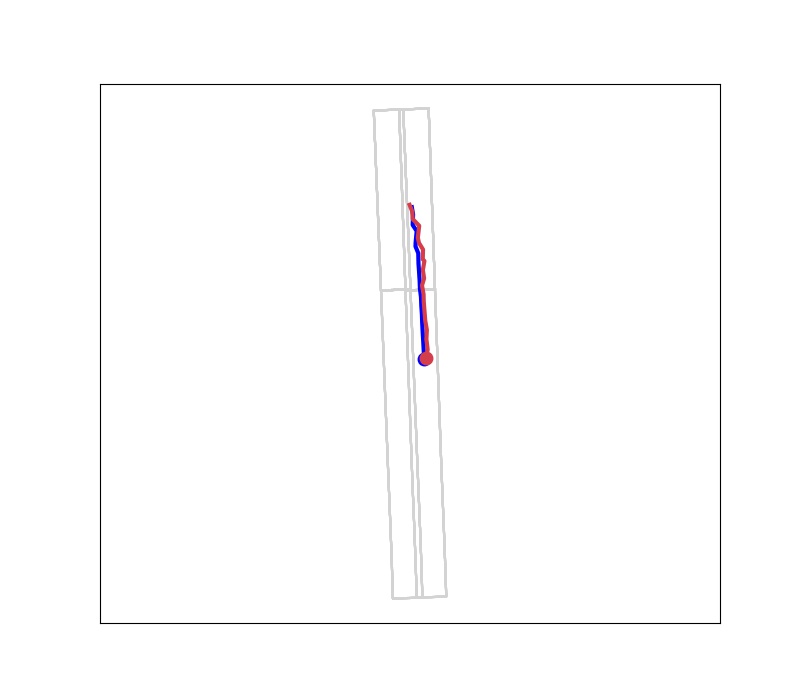}
\\
\includegraphics[width=1.55in]{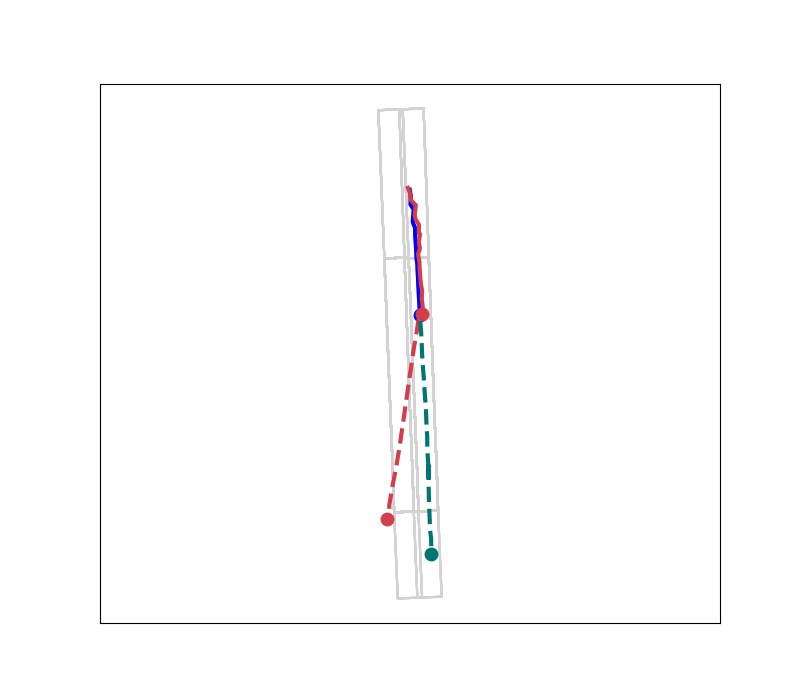}
\end{minipage}%
}%
\subfigure[]{
\begin{minipage}[t]{0.22\linewidth}
\centering
\includegraphics[width=1.55in]{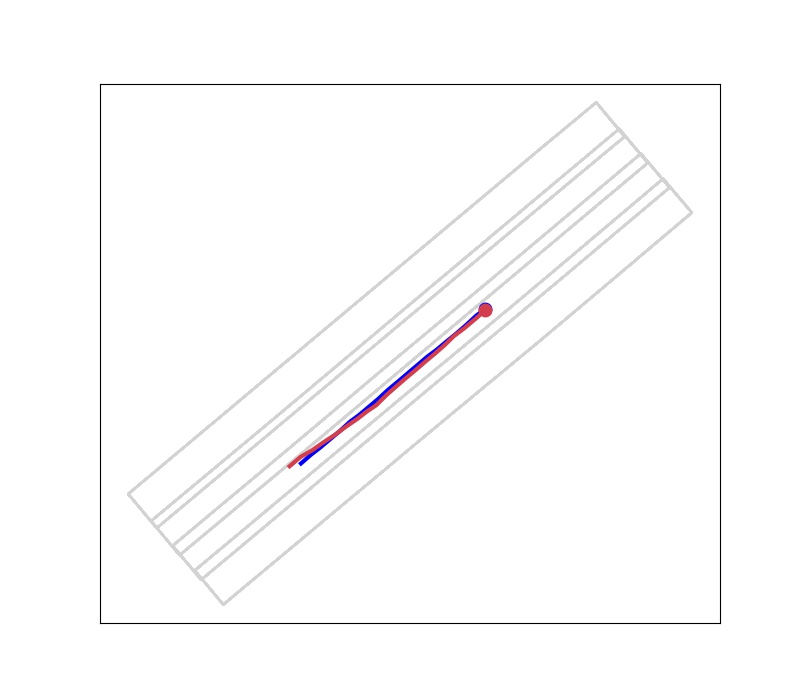}
\\
\includegraphics[width=1.55in]{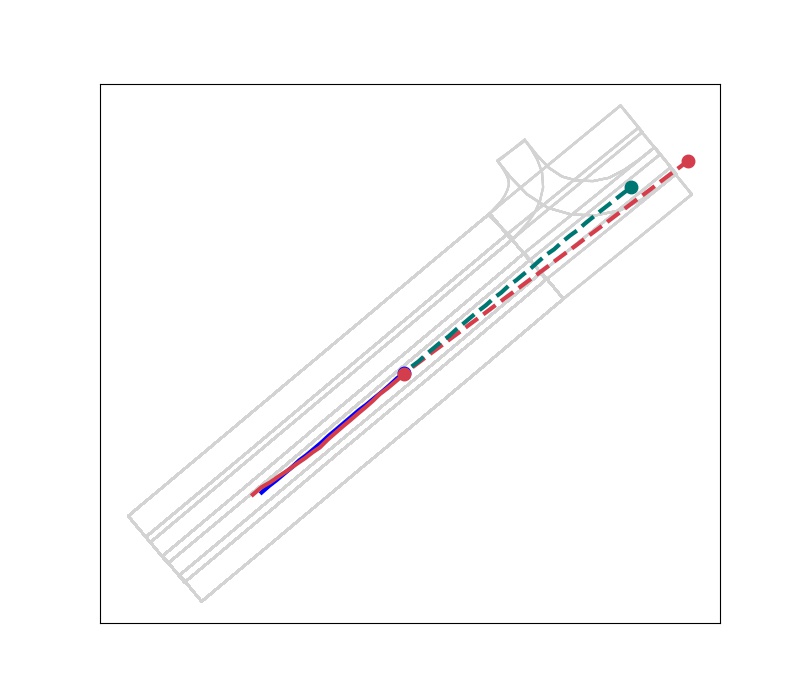}
\end{minipage}%
}%

\centering
\caption{Adversarial trajectories generated by different types of attacks. The figures in the top row are adversarial (red line) and benign (blue line) input history trajectories for prediction and they look very close to human eyes.
The figures in the bottom row show the corresponding attacked future trajectory prediction (red dashed line) and the ground truth trajectory in the benign case (green dashed line). The differences between the two are clearly visible, showing the effectiveness of the attacks. Figure (a) is the ADE attack, which will randomly lead to maximum average deviation. Figure (b) is the lateral attack, which mainly leads the vehicle to deviate to the left or right. Figure (c) is the longitudinal attack, which will mainly lead to longitudinal deviation.} 
\label{fig:adv_trajs}
\end{figure*}

Different from image classification, trajectory prediction has no class labels, but it has directional information in the context, such as moving forward, turning or changing lane to the right. Therefore, the attacker can conduct targeted attacks besides the random ones. In Fig.~\ref{fig:adv_trajs}, three types of adversarial attacks are presented and we observe that they can cause significant and directional errors.   \cite{zhang2022adversarial} proposes directional error metrics for the optimization of targeted attacks, as shown in Eq.~\eqref{eq:metrics}:
\begin{equation}
D(\alpha, R)=\left(p_{\alpha}-s_{\alpha}\right)^{T} \cdot R\left(s_{\alpha+1}, s_{\alpha}\right), \label{eq:metrics}
\end{equation}
where $\alpha$ denotes the time frame ID. $p$ and $s$ are two-dimensional vectors representing predicted and ground-truth vehicle locations, respectively.  $R$ is a function generating the unit vector to a specific direction (lateral or longitudinal).
The longitudinal direction is approximated as the vector defined by the adjacent two waypoints of ground truth $s_{\alpha+1}^{n}-s_{\alpha}^{n}$.

In addition to the directional attack, a random attack can be designed to maximize the Average Displacement Error (ADE),  which is the average of the root mean squared error between the predicted waypoints and the ground-truth trajectory waypoints. 

Hard constraints are also applied to the maximum deviation of waypoints during the optimization so as to make the adversarial history trajectory physically feasible and not perform unrealistic behaviors.

\begin{table}[t]
\centering
\caption{Trajectory prediction errors (in meters) in benign or adversarial scenarios under different types of attacks. }   
\resizebox{.85\columnwidth}{!}{
\label{tab:pre-attack}
\begin{tabular}{c|c|c|c}
\hline
\multirow{2}*{Model}    & ADE  & Lateral  & Longitudinal  \\
\cline{2-4}
&benign/attack &benign/attack &benign/attack \\
\hline
LaneGCN \cite{liang2020learning} &$1.32/5.17$ &$-0.01/1.58$ & $-0.25/3.33$ \\ \hline
Trajectron++\cite{salzmann2020trajectron++} & $2.69/6.81$  &$0.107/2.25$  &-0.526 / 3.79 \\ \hline

\end{tabular}
}
\end{table}

Table~\ref{tab:pre-attack} quantitatively shows that the predicted trajectories can be seriously deviated by any of the three attack types. Generally, in the US, 0.3m lateral deviation is enough to invade adjacent lanes on local roads~\cite{sato2021dirty,hancock2013policy}. In addition, the final displacement error (the distance between the last points of predicted and ground-truth trajectories) is around two to three times larger than the average error shown here. Therefore, the attacker can apply either random or directional attacks that will greatly challenge the defense methods, especially on the generalization performance. 

\cite{zhang2022adversarial,caorobust} propose to utilize approaches such as smoothing and adversarial training on the conditional variational encoder (CVAE) architectures to mitigate the impact of such adversarial attacks. However, they do not explicitly address driving semantics and robustness generalization, which are critical for many real road scenarios.

\subsection{Adversarial Training Methods and Robust Generalization}

Adversarial training~\cite{madry2017towards, maini2020adversarial, schott2018towards} is shown to be one of the most effective approaches to improve the robustness of DNN models. In practice, the PGD attack is commonly deployed for evaluation because of its strong attack ability in white-box settings and the work in~\cite{madry2017towards} formulates the adversarial training as a min-max problem. \cite{gowal2021improving,rebuffi2021fixing,sehwag2021robust} show that synthesized data can significantly boost robustness for the image classification task. Thus, we select similar methods as baselines for comparison in our experiments. 

Recent research shows that a specific type of adversarial attack is not sufficient to represent the diverse space of adversarial examples and many adversarially trained models are only robust to specific attacks. This does limit the application of adversarial training in practice, especially for trajectory regression tasks due to its nature of long-tailed distribution. The works in~\cite{schmidt2018adversarially,zhang2021towards,song2019robust} try to explain and improve the robust generalization in various perspectives such as sample complexity and latent feature representation. In this work, we demonstrate that robust generalization for trajectory prediction can be enhanced by explicitly introducing disentangled and semantic features in the latent space. 

\subsection{Adversarial Autoencoder Architecture}
The adversarial autoencoder (AAE) is a variant of the variational autoencoders (VAE)~\cite{kingma2013auto,kingma2019introduction}, which provides a principled method for jointly learning deep latent-variable models and corresponding inference models using stochastic gradient descent. 
AAE imposes various distributions on the latent vector by utilizing adversarial learning instead of KL divergence. Due to the flexibility of adversarial learning, AAE is superior to VAE in terms of imposing complicated distributions over latent space. For adversarial robustness, the works in~\cite{mathieu2019disentangling, willetts2019improving} demonstrate that the disentangled latent representations produce VAEs that are more robust to adversarial attacks. In our work, we design an AAE-based architecture that can be added after the feature extractor of prediction modules. We utilize this architecture to model diverse semantic features and enhance disentanglement in the latent space.

 
\section{Our Proposed  Adversarial Training Method for Trajectory Prediction}
\label{sec:method}
\subsection{Domain Knowledge-guided Semi-supervised Architecture}
\label{sec:model}
\subsubsection{Overall Design}
\begin{figure*}[!ht]
    \centering
    \includegraphics[width=1.7\columnwidth]{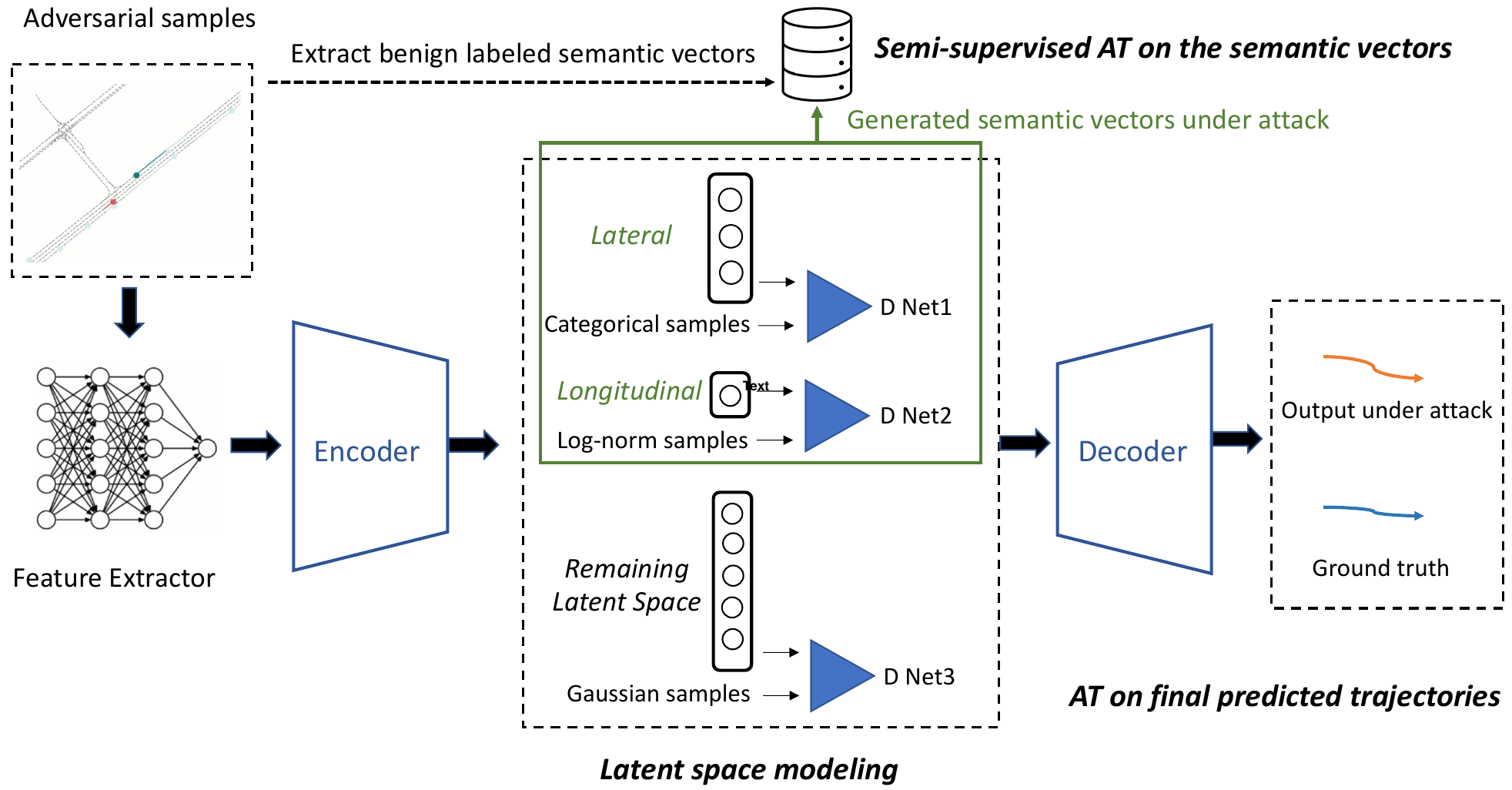}
    \caption{The overall adversarial training pipeline for the proposed semi-supervised semantics-guided architecture. We leverage the feature extractor in~\cite{liang2020learning}. In the latent space, the D nets represent the discriminator that models the latent distributions by distinguishing the samples of real distribution from the predicted ones. When the semantic labels are available in the scenarios, they are used to adversarially train the semantic vectors (parts in green). Finally, we conduct adversarial training on attacked trajectories. }
    \label{fig:arch}
\end{figure*}

Unlike adversarial training for image classification tasks, there are no class labels to guide the training for trajectory prediction, and the attack patterns in the trajectory regression are more random. We thus introduce domain knowledge to facilitate the modeling of the semantic information in both benign and adversarial cases, based on an AAE architecture. The model learns the directional semantic latent vector in a semi-supervised way because the ground truths are only available for limited scenarios but their distributions can be derived from domain knowledge and statistics.  Therefore, our latent space modeling contains two levels -- unsupervised distribution modeling and semi-supervised learning when labels are available.

The architecture of our proposed model is shown in Fig.~\ref{fig:arch}. The feature extractor~\cite{liang2020learning} utilizes a one-dimensional dilated convolutional neural network to obtain the embedding of the time-series trajectory and uses a graph neural network to model the lane context and interaction between objects. The encoder maps high-dimensional features to the semantics-guided latent space with distribution regularization and semi-supervised training. In particular, the latent space is divided into three parts: longitudinal features $z_{lon}$, following one-dimensional log-normal distribution, lateral features $z_{lat}$, following three-dimensional categorical distribution and remaining features, following Gaussian distribution~\cite{jiao2022tae}. Finally, the decoder maps semantic vectors along with other disentangled latent vectors to the future trajectories. Note that we develop the AAE instead of traditional VAE architecture to model these different and complex distributions.

In the attacked scenarios, the impact of attacks will be decomposed to different latent vectors and the attack patterns will be explicitly modeled by the semantic features. Let us take the lateral directional vector as an example. If the attack is not targeted at the lateral dimension, the encoder will decompose the attack effect into other vectors and the mapping for lateral direction will remain stable. Otherwise, if the attack causes errors in the lateral vector, the feature extractor and encoder will be adversarial trained on the label of the direction and the corresponding mapping from the latent distribution to the final trajectories will also be updated.

Compared to existing trajectory prediction works that only apply adversarial training on the final trajectory waypoints, our proposed method is designed to capture semantic features in the latent space, and it can benefit the adversarial robustness and its generalization in various aspects: 

\begin{itemize}

\item{The architecture maps the high-dimensional features into disentangled latent space that can decompose the attacks into different orthogonal patterns. Such disentanglement will boost the adversarial robustness~\cite{willetts2019improving}}.

\item{The semantic vectors provide more context information and interpretable labels for the adversarial training and the semi-supervised learning makes the adversarial training process more efficient and easier to generalize to the unseen scenarios.}

\item{The encoder-decoder architecture can filter high-frequency noise by dimension reduction.}

\end{itemize}

The detailed design and optimization of the architecture are described in the following.



\subsubsection{Semi-supervised Semantic Features Modeling}
\label{sec:semantics-modeling}

To model the high-level semantic information in the trajectory prediction task, it is natural to decompose the trajectories into two dimensions -- longitudinal and lateral directions. And we want to utilize domain knowledge to guide the modeling by providing representative metrics and prior distributions. In addition to the semantic features, the architecture also maps high-dimensional information into latent vectors in Gaussian distributions to represent other low-level and more random features. We will mainly explain the semantic feature modeling below.

In the longitudinal direction, speed and acceleration are often used to model the vehicle's dynamic but their values are always changing and do not contain enough semantic information. In our model, we apply the \emph{time headway} to effectively extract the longitudinal feature, which measures the time difference between two successive vehicles when they cross a given point. The time headway represents a relatively stable behavior pattern for a certain agent and takes interaction with other vehicles into consideration. Recent works~\cite{yu2018human, liu2020impact} also use time headway as a measurement of aggressiveness in specific scenarios such as lane changing or merging. In our proposed architecture, the model will represent the time headway as a one-dimensional vector in the latent space. In both benign and adversarial cases, the encoder will be trained by the regularization loss to force the longitudinal feature to a certain distribution that the time headway follows in statistics. Prior works~\cite{ha2012time} in the transportation domain show that the time headway in urban scenarios can be best described with the Log-normal distribution. We verify this and estimate the parameters in the Argoverse 1 motion forecasting dataset~\cite{chang2019argoverse}. The longitudinal feature follows the distribution shown in Eq.\eqref{eq:log-normal}:

\begin{equation}
    f(x)=\frac{1}{x \sigma \sqrt{2 \pi}} e^{-\frac{1}{2}\left(\frac{\ln (x)-\mu}{\sigma}\right)^{2}}\label{eq:log-normal},
\end{equation}
where $\mu$ is the location parameter and estimated as 0.0682. $\sigma$ is the scale parameter and estimated as 0.647. We can explicitly obtain the true time headway values for semi-supervised training when there is observable interaction between the attacked target and the front vehicle. We consider the semi-supervised longitudinal feature encoding as a regression problem and optimize it by minimizing the mean square error.

For the lateral directional features, we represent them by three simple but effective classes: moving forward, turning/changing lane to the left, and turning/changing lane to the right. These three intentions are discrete by nature and we model them with the categorical distribution. In the adversarial training process, only vehicles with clear intentions in a long enough time frame will be labeled and we utilize the cross-entropy to optimize this classification task.

For all the semantic and Gaussian latent variables, they are regularized to the target distribution by the adversarial generation loss in Eq.~\eqref{eq:adv-loss}. The discriminators are trained to maximize the log probability of real latent samples $s$ and the log of the inverse probability for fake latent samples, as in Eq.~\eqref{eq:dis}:
\begin{align}
Loss_{G}(x) &= \frac{1}{m} \sum_{i=1}^{m} \log \left(1-D_i\left(G\left(x\right)\right)\right)\label{eq:adv-loss}, \\
Loss_{D_i}(x,s) &= \log D_i\left(s_i\right)+\log \left(1-D_i\left(G\left(x\right)\right)\right)\label{eq:dis},
\end{align}
where $x$ is the high-dimensional features and $m$ is the number of different kinds of latent vectors. $G$ and $D$ are encoders and distribution discriminators, respectively.




\subsection{Adversarial Training Process for Trajectory Prediction}
\label{sec:AT}
\subsubsection{Adversarial Training Algorithm}

For each sample, we utilize the PGD attack to generate the adversarial trajectory only for the target vehicle and keep other surrounding vehicles' original trajectories, by which, the adversarial attack's impact on the whole scenario is constrained. If the error between the prediction under attack and the ground truth is greater than a threshold, we consider the attack a successful one and conduct adversarial training on this sample. Since the perturbation is very small, we consider the real future trajectory as the ground truth $y_{i}$ for the adversarial training and optimize the whole pipeline with L1-smooth loss in Eq.~\eqref{eq:l1-smooth}. This is a general but somewhat naive adversarial training process. 
\begin{equation}
{Loss}_{traj}\left(y_{i}, \hat{y}_{i}\right)= \begin{cases}0.5\left(y_{i}-\hat{y}_{i}\right)^{2} & \text { if }\left\|y_{i}-\hat{y}_{i}\right\|<1, \\ \left\|y_{i}-\hat{y}_{i}\right\|-0.5 & \text { otherwise }.\end{cases} \label{eq:l1-smooth}
\end{equation}

Thus, to further facilitate the adversarial training, we exploit semantic features and their corresponding labels in our proposed architecture. The encoder is optimized to minimize the mean square error of longitudinal features and the cross entropy of lateral features between the ground truths and predictions in the latent space.  The semi-supervised loss function is shown as follows:
\begin{equation}
Loss_{Semi}(z,g) =  -\sum_{i=1}^3g_{lat}\log z_{lat} + (g_{lon} - z_{lon})^2, \label{eq:semi-loss}
\end{equation}
where $z$ represents the predicted semantic vectors when under attack and $g$ represents the ground truth in the benign scenarios.

Moreover, we further adapt the adversarial training process with lateral semantic vectors because the lateral directional prediction can be regarded as a classification problem with clear behavior meaning. When the adversarial example leads to the wrong classification of lateral behavior, we will set higher weights of the semi-supervised loss for the adversarial training. In this way, our model will first guarantee the correctness of high-level semantic prediction and then tune the regression error, which could help avoid significant adversarial deviation and enhance the generalization performance.





\begin{algorithm}[htbp]
\caption{Our Adversarial Training Pipeline}
\label{alg:pipeline}
\begin{algorithmic}[1]
\STATE\textbf{Initialize:} feature extractor $F$, AAE encoder $G$, decoder $R$, discriminator $D_{i}$, target distribution $p_{i}$, $i$ = 1,2,3, adversarial example generator $Adv$.

\STATE\textbf{Input:} past trajectories $t$, future trajectories $y$ , map context $c$, prediction model $m$.

\FOR{each sample}
\STATE Generate adversarial trajectory for target vehicle $t_{adv} = Adv(t,y,m,c)$.
\STATE Input $t_{adv}$ to the predictor. $\hat y', \hat z' = m(t_{adv},c)$, where $z$ =$\{z_{lat},z_{lon},z_{gaussian}\}$.
\IF{$err(\hat y', y) > Threshold$ }
    \IF {$argmax(z_{lat}) \neq argmax(\hat {z_{lat}}')$}
    \STATE Update model $m$ with higher weights of semi-supervised learning loss $Loss_{semi}$ in Eq.~\eqref{eq:semi-loss}.
    \STATE Update model $m$ with $Loss_D$ in Eq.~\eqref{eq:dis}, $Loss_{G}$ in Eq.~\eqref{eq:adv-loss}, and $Loss_{traj}$ in Eq.~\eqref{eq:l1-smooth}.
    \ELSE
    \STATE Update model $m$ with $Loss_D$ in Eq.~\eqref{eq:dis}, $Loss_{G}$ in Eq.~\eqref{eq:adv-loss}, $Loss_{semi}$ in Eq.~\eqref{eq:semi-loss}, and $Loss_{traj}$ in Eq.~\eqref{eq:l1-smooth}. 
    \ENDIF
\ENDIF

\ENDFOR

\end{algorithmic}
\end{algorithm}

\subsubsection{Balance Accuracy and Robustness} 

In our preliminary experiments, we notice a trade-off between standard accuracy and adversarial robustness. A similar phenomenon has been observed in classification tasks~\cite{zhang2019theoretically,raghunathan2020understanding}. Methods such as TRADES~\cite{zhang2019theoretically}, robust self-training~\cite{carmon2019unlabeled} and MixUp~\cite{zhang2017mixup, archambault2019mixup} have been proposed to balance such trade-off. 
However, there are few methods that can be applied to trajectory prediction because such time-series regression problems have no class labels and are more sensitive to errors introduced by augmented data. In this work, we utilize the MixUp~\cite{zhang2017mixup} technique to mix the adversarial scenarios and benign scenarios in the adversarial training process. The experiments demonstrate that a balance between adversarial robustness and standard accuracy can be achieved in trajectory prediction, as shown later.

\section{Experimental Results}
\label{sec:exp}
\subsection{Experiment Setup}

\subsubsection{Dataset} We train and evaluate different defense methods using three popular benchmarks -- Argoverse 1~\cite{chang2019argoverse}, Argoverse 2 ~\cite{wilson2argoverse}, and ApolloScpae datasets~\cite{huang2018apolloscape}. The datasets contain more than 250k real driving scenarios in different cities, such as Miami and Pittsburgh. For Argoverse 1 and Argoverse 2, each scenario consists of a road graph and multiple agents' trajectories sampled at a frequency of 10Hz. We choose 20 waypoints as the history trajectory and the models will predict 30 waypoints in the future. Scenarios in Apolloscape are simpler. They have no maps but 6 waypoints for both history and future trajectories.  

\begin{table*}[h]
\centering
\caption{Comparison of different defense methods when under various attacks and in the benign case. For the adversarial training methods, we calculate the mean error of models adversarially trained on different attacks.}   
\resizebox{2\columnwidth}{!}{
\label{tab:defense_comp}
\begin{tabular}{|c|c|c|c|c|c|c|c|c|c|c|c|c|c|c|}
\hline
\multirow{2}*{Methods}   & \multicolumn{3}{|c|}{ADE Attack} & \multicolumn{3}{|c|}{Lateral Attack}& \multicolumn{3}{|c|}{Lonngitudinal Attack} &  \multicolumn{3}{|c|}{Benign}\\
\cline{2-13}

&\multicolumn{3}{|c|}{ADE/m} &\multicolumn{3}{|c|}{Lateral Error/m} &\multicolumn{3}{|c|}{Longitudinal Error/m} & \multicolumn{3}{|c|}{ADE/m}\\
\hline

Dataset & Argo1 & Argo2 &Apollo & Argo1 & Argo2 &Apollo & Argo1 & Argo2 &Apollo  & Argo1 & Argo2 &Apollo\\ \hhline{|=|=|=|=|=|=|=|=|=|=|=|=|=|} \hline

Original Model  &$5.17$ &$4.52$ &$3.93$&$1.66$& $1.27$&$1.54$ &$3.78$ &$3.17$&$3.42$ &\textbf{1.43} & $\textbf{0.79}$ & $\textbf{1.78}$\\ \hline 
Train-time Smo.~\cite{zhang2022adversarial} & $4.67$ &$4.32$&$3.86$ &$1.45$&$1.27$&$1.49$  &$3.23$ &$3.00$ & $3.41$&$ 1.50$ & $0.83$ & $2.11$ \\ \hline 
Test-time Smo.~\cite{zhang2022adversarial} & $4.32$ & $3.51$ & $3.49$ &$ 0.75$ & $0.89$ & $1.09$ &$3.54$ & $2.42$ & $3.05$ &$1.68 $ & $1.23$ & $2.07$ \\ \hline  
Heuristic Aug.~\cite{rebuffi2021fixing} & $4.62 $ &$3.52$ &$3.84$ &$ 1.09$ & $1.21$ &$1.34$ &$ 3.24$& $4.24$ &$3.21$ & $1.54 $ & $0.97$ & $2.01$ \\ \hline 
Data-driven Aug.~\cite{rebuffi2021fixing} & $4.53 $ &$4.43$ &$3.53$ &$ 0.74$  & $0.78$ & $0.54$ &$ 2.73$ & $2.61$ & $3.02$ &$2.50$ & $1.99$ & $2.24$ \\ \hline
Standard AT & $3.79$ & $3.67$ & $3.68$ &$0.67$& $0.64$ & $1.15$ &$2.44$ & $1.89$ & $3.05$ &$1.67 $ & $1.06$ & $1.87$ \\ \hhline{|=|=|=|=|=|=|=|=|=|=|=|=|=|}
SSAT (ours) & $\textbf{3.51}$ & $\textbf{2.67}$ & $\textbf{2.90}$ &$\textbf{0.61}$ & $\textbf{0.53}$ & $\textbf{0.41}$ &$\textbf{1.76}$ & $\textbf{1.44}$ & $\textbf{1.26}$ &$ 1.75$ & $1.20$ & $1.87$\\ \hline 
Mixup-SSAT (ours) & $3.73$ & $3.33$ & $3.17$ &$0.72$ & $0.51$ & $0.53$ &$2.13$ &  $1.63$  & $1.35$  &$ 1.64$ & $1.07$ & $1.86$\\ \hline 

\end{tabular}
}
\end{table*}

\subsubsection{Attack Settings}
In the experiments, we study three different types of attacks~\cite{zhang2022adversarial} to the vehicle trajectory prediction algorithms -- lateral directional attack (shift to the right), longitudinal directional attack (shift forward), and ADE attack (deviate randomly).  More details of them are in Section~\ref{sec:pre-att}. The trajectory prediction models are adversarially trained on the three types of attacks, respectively. We constrain the maximum deviation between the attacked and the benign input trajectories to be 1 meter. 

\subsubsection{Training Settings}
Since our architecture is an encoder-decoder module that can be combined with different feature extractors, we first fine-tune the model on the benign data. In the experiments, we use the feature extractor from LaneGCN~\cite{liang2020learning}, an attention-based graph neural network. We notice that the AAE architecture introduces a slight accuracy drop on the benign data, mainly due to the dimension reduction. For adversarial training, we train prediction models on adversarial samples generated from scenarios in datasets. 

\subsection{Experimental Results and Analysis}
In this section, we conduct experiments with various defense methods under different patterns of attacks, including our semi-supervised semantics-guided (SSAT) method and a Mixup-SSAT method that combines SSAT with the MixUp technique for balancing standard accuracy and adversarial robustness. In the following, we first compare the average robustness improvement among various defense methods, which demonstrates the advantage of SSAT in improving robustness under various types of attacks and the effectiveness of Mixup-SSAT in balancing robustness and accuracy. Then, we show that SSAT can significantly enhance the robust generalization to unseen types of attacks. In addition, we evaluate an unsupervised version of SSAT to explicitly show how the semi-supervised semantic-guided latent space modeling can boost the adversarial robustness, which also serves as an ablation study. 

\subsubsection{Effectiveness of Our SSAT Methods}
We compare \emph{our SSAT and Mixup-SSAT methods with the original model and five different defense methods}, including train-time smoothing~\cite{zhang2022adversarial}, test-time smoothing~\cite{zhang2022adversarial}, heuristic data augmentation~\cite{rebuffi2021fixing}, data-driven augmentation~\cite{rebuffi2021fixing}, and the standard adversarial training (Standard AT). All the models share the same feature extractor LaneGCN in this setting. 
Note that we compare to the two data augmentation methods as they are effective for image classification tasks~\cite{gowal2021improving,rebuffi2021fixing,sehwag2021robust}.
For the data-driven augmentation, we design an additional decoder to augment input trajectories and it can generate more inputs by adding Gaussian noises to the latent vectors of the real inputs.  For the heuristic augmentation, we simply add random perturbation to the benign inputs, with the same constraints of maximum deviation. 

Table~\ref{tab:defense_comp} shows the prediction errors of different methods when under various types of attacks and in the benign case (note that when under lateral and longitudinal attacks, we measure lateral and longitudinal errors). We can see that \textbf{our SSAT method significantly outperforms all other defense methods in improving the robustness of trajectory prediction}. Compared with the original model, SSAT can reduce the prediction error by $32\%-73\%$ when under different types of attacks.
\begin{figure*}[htbp]

\centering

\subfigure[]{
\begin{minipage}[t]{0.32\linewidth}
\centering
\includegraphics[width=1.8in]{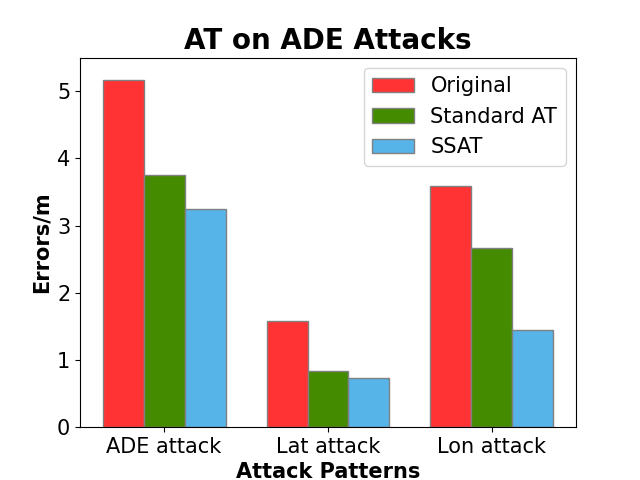}
\end{minipage}%
}%
\subfigure[]{
\begin{minipage}[t]{0.32\linewidth}
\centering
\includegraphics[width=1.8in]{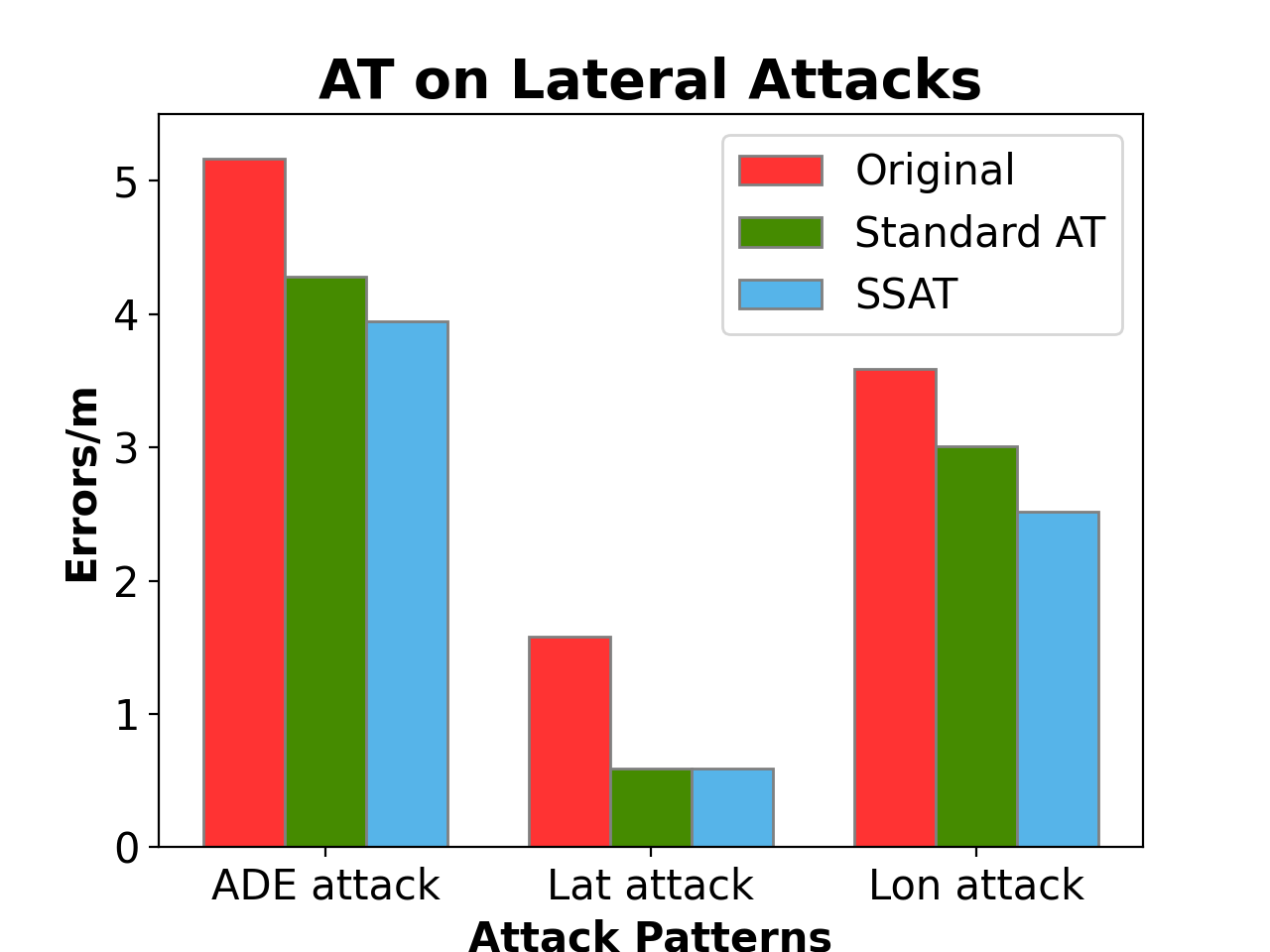}
\end{minipage}%
}%
\subfigure[]{
\begin{minipage}[t]{0.32\linewidth}
\centering
\includegraphics[width=1.8in]{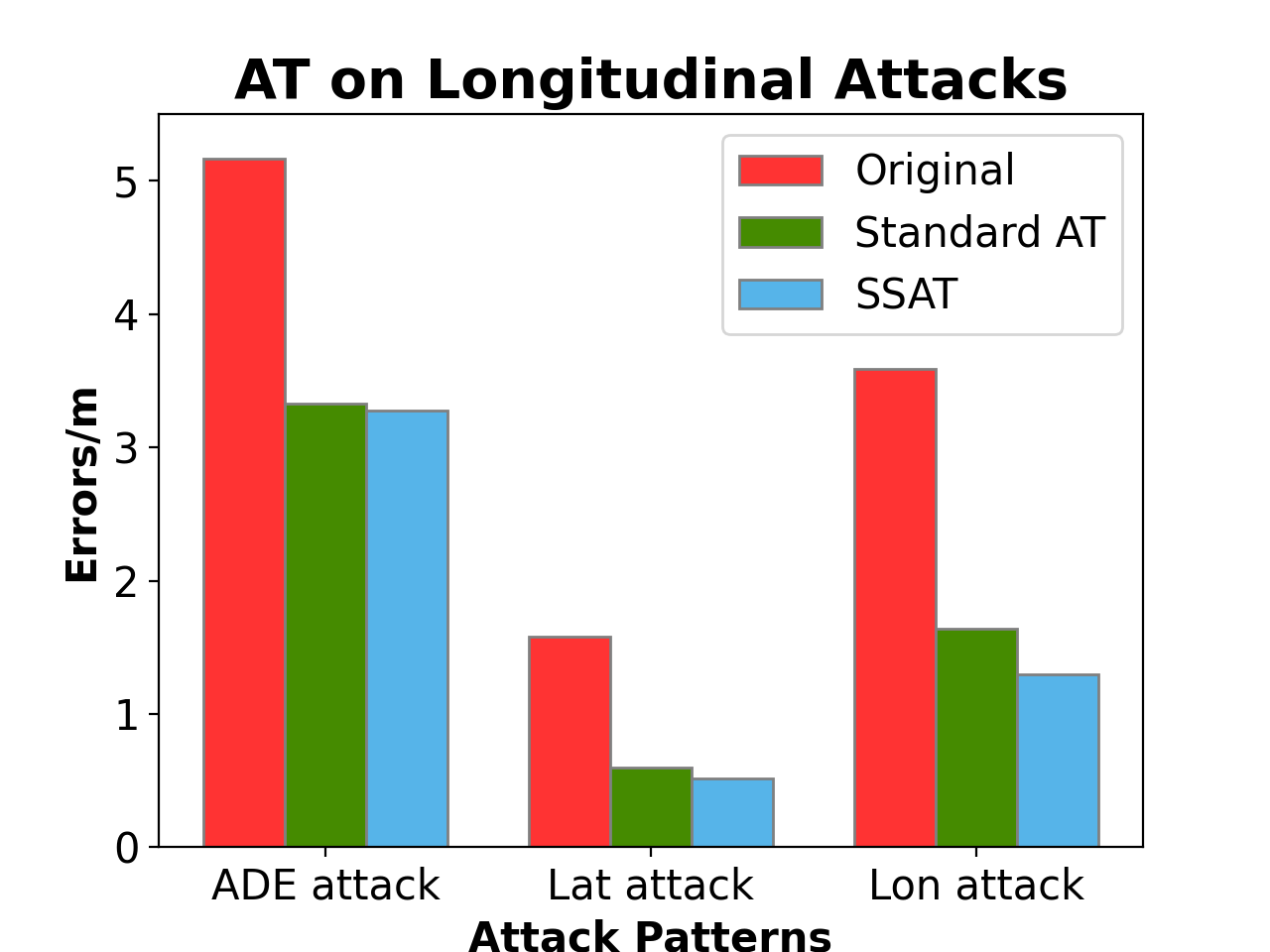}
\end{minipage}%
}%

\caption{Adversarial robustness comparisons of the original model, the standard adversarial training on final trajectories (Standard AT), and the SSAT method proposed by us. Figures (a), (b), (c) show results from adversarial training (AT) that targets ADE attacks, lateral attacks, and longitudinal attacks, respectively.}
\label{fig:AT-bar}
\vspace{-0.5cm}
\end{figure*}
Moreover, while our SSAT method improves robustness, we also observe a drop in standard accuracy in the benign case. \textbf{Mixup-SSAT enables effective trade-off between these two objectives} (i.e., better performance than SSAT in benign cases at the expense of worse performance under attacks), by setting the mixup ratio between adversarial and benign examples to different values (the results in Table~\ref{tab:defense_comp} are based on a mixup ratio of 2). 


We also notice that both data-driven and heuristic data augmentation methods offer very limited improvement over the original model. This is likely due to the challenge of regression tasks with rich context, which makes augmentation methods perform poorer than they do for image classification tasks.




\subsubsection{Effectiveness of SSAT in Robust Generalization on Different Types of Attacks}

We observe that there is an adversarial robust generalization gap when the training and test are under different types of attacks. The comparisons in Tables~\ref{tab:at-ade},~\ref{tab:at-lat},~\ref{tab:at-lon} show that \textbf{our SSAT method is better generalized to unseen types of attacks}, compared to the standard adversarial training. In every training scenario, our method is more robust to the various unseen patterns of attacks. 
For instance, Table~\ref{tab:at-ade} shows that when applying SSAT to train under the random ADE attack, its results outperform other models on all seen (i.e., ADE) and unseen (i.e., lateral and longitudinal) types of attacks, which demonstrates that our SSAT method can effectively decompose and learn semantic features from random ADE attacks. Tables~\ref{tab:at-lat} and~\ref{tab:at-lon} show similar trends, where our SSAT methods are better at defending against unseen attacks and mitigating overfitting on specific patterns of attacks. Fig.~\ref{fig:AT-bar} further visualizes the results from these three tables for the original model, the standard adversarial training (Standard AT), and SSAT.  
\begin{table}[H]
\centering
\caption{Comparison of different methods when adversarial training is conducted for \textbf{ADE} attack, and tested on ADE, lateral, and longitudinal attacks.} 

\label{tab:at-ade}
\resizebox{.85\columnwidth}{!}{
\begin{tabular}{c|c|c|c}

\hline
\multirow{2}*{Methods}  
& \multicolumn{3}{c}{ Prediction Error/m, w/o AT $\rightarrow$ with AT }  \\
\cline{2-4}
&ADE Attack&Lat Attack &Lon Attack \\
\hline
Standard AT   &$5.17\rightarrow3.76$ &$1.58\rightarrow0.83$ & $3.59\rightarrow2.66$\\ \hline 
SSAT & $5.16\rightarrow\textbf{3.24}$  &$1.66\rightarrow\textbf{0.73}$  &$3.78\rightarrow\textbf{1.45}$\\ \hline 
Unsup-SSAT   &$5.16\rightarrow3.48$ &$1.66\rightarrow0.74$ &$3.78\rightarrow1.49$ \\ \hline

\end{tabular}
}
\end{table}

\subsubsection{Impact from Latent Space Modeling}
We also conduct adversarial training that only has regularization on the latent distributions but without supervision on latent vectors. We name it Unsup-SSAT. The comparisons between the standard adversarial training (Standard AT) and Unsup-SSAT in Tables~\ref{tab:at-ade},~\ref{tab:at-lat}, and~\ref{tab:at-lon} demonstrate that even without labels, the partial disentanglement and distribution modeling in Unsup-SSAT will benefit the adversarial training on trajectory prediction and outperform the baseline Standard AT in most cases. However, when compared with SSAT, we find that the adversarial robustness will be further improved with the extra labels in the semi-supervised phase (in practice, we often have access to those labels). 


\begin{table}[H]
\centering
\caption{Comparison of different methods when adversarial training is conducted for \textbf{lateral} attack, and tested on ADE, lateral, and longitudinal attacks. }   
\label{tab:at-lat}
\resizebox{.85\columnwidth}{!}{
\begin{tabular}{c|c|c|c}
\hline
\multirow{2}*{Methods}   & \multicolumn{3}{c}{ Prediction Error/m, w/o AT $\rightarrow$ with AT }  \\
\cline{2-4}
&ADE Attack&Lat Attack &Lon Attack \\
\hline
Standard AT&$5.17\rightarrow4.28$ &$1.58\rightarrow\textbf{0.59}$ & $3.59\rightarrow3.01$ \\ \hline
SSAT &$5.16\rightarrow\textbf{4.02}$ &$1.66\rightarrow\textbf{0.59}$&$3.78\rightarrow\textbf{2.52}$ \\ \hline
Unsup-SSAT   &$5.16\rightarrow4.11$ &$1.66\rightarrow0.69$   &$3.78\rightarrow2.54$ \\ \hline

\end{tabular}
}
\end{table}

\begin{table}[H]
\centering
\caption{Comparison of different methods when adversarial training is conducted for \textbf{longitudinal} attack, and tested on ADE, lateral, and longitudinal attacks.
}   

\label{tab:at-lon}
\resizebox{.85\columnwidth}{!}{
\begin{tabular}{c|c|c|c}
\hline
\multirow{2}*{Methods}   & \multicolumn{3}{c}{ Prediction Error/m, w/o AT $\rightarrow$ with AT}  \\
\cline{2-4}
&ADE Attack&Lat Attack &Lon Attack \\
\hline
Standard AT   &$5.17\rightarrow3.33$ &$1.58\rightarrow0.60$ &$3.59\rightarrow1.64$ \\ \hline 
SSAT & $5.16\rightarrow\textbf{3.28}$  &$1.66\rightarrow\textbf{0.52}$  &$3.78\rightarrow\textbf{1.30}$ \\ \hline 
Unsup-SSAT   &$5.16\rightarrow3.40$ &$1.66\rightarrow0.61$ & $3.78\rightarrow1.49$ \\ \hline 

\end{tabular}
}
\end{table}

\section{Conclusion}
In this work, we propose an adversarial training method for trajectory prediction.
To tackle the challenge of random inputs with rich context, diverse types of attacks, and lack of class labels, we develop a novel AAE architecture that exploits the disentanglement and semantic features for enhancing model robustness and its generalization. Our proposed SSAT method significantly outperforms a number of baselines from the literature, reducing the prediction errors under attacks by up to $32\%-73\%$ when compared with the original prediction models. Our method is also shown to be effective for defending against unseen attacks. 

\bibliographystyle{plain} 
\bibliography{references}

\end{document}